%% file: acl.tex
\pdfoutput=1

\documentclass[11pt]{article}

\usepackage[]{acl}

\usepackage{times}
\usepackage{latexsym}
\usepackage{graphicx}
\usepackage[size=tiny]{todonotes}
\usepackage[T1]{fontenc}
\usepackage[utf8]{inputenc}
\usepackage{microtype}
\usepackage{amsmath}
\usepackage{amssymb}
\usepackage{booktabs}
\usepackage{enumitem}
\usepackage{subcaption}

\title{BMX: Boosting Natural Language Generation Metrics with Explainability}

\author{Christoph Leiter$^1$, Hoa Nguyen$^2$, Steffen Eger$^1$ \\
$^1$ Natural Language Learning Group (NLLG)\\ \url{https://nl2g.github.io/}\\
$^1$ University of Mannheim, $^2$ TU Darmstadt\\
\texttt{\{christoph.leiter,steffen.eger\}@uni-mannheim.de}}

\newcommand*\red[1]{\textcolor{black}{#1}}
\newcommand*\tablegreen[1]{\textcolor{blue}{#1}}

\newcommand{\se}[1]{\textcolor{black}{#1}}
\newcommand{\senew}[1]{\textcolor{black}{#1}}
\newcommand{\cl}[1]{\textcolor{black}{#1}}
\newcommand{\cln}[1]{\textcolor{black}{#1}}
\newcommand{\cli}[1]{\textcolor{black}{#1}}
\newcommand{\rev}[1]{\textcolor{black}{#1}}

\newcommand\smallE{
  \mathchoice
    {{\scriptstyle\mathcal{E}}}
    {{\scriptstyle\mathcal{E}}}
    {{\scriptscriptstyle\mathcal{E}}}
    {\scalebox{.7}{$\scriptscriptstyle\mathcal{E}$}}
  }

\begin{document}
\maketitle
\begin{abstract}
State-of-the-art natural language generation evaluation metrics are based on black-box language models. Hence, recent works consider their explainability with the goals of better understandability for humans and better metric analysis, including failure cases. In contrast, \cli{our proposed method BMX: Boosting Natural Language Generation Metrics with explainability} explicitly leverages explanations to \cli{boost} the metrics' performance. In particular, we perceive feature importance explanations as word-level scores, which we convert, via power means, into a segment-level score. We then combine this segment-level score with the original metric to obtain a better metric. 
\cln{Our tests show improvements for multiple metrics
\senew{across MT and summarization datasets.}
While improvements \cli{in} machine translation are small, they are strong for summarization}. Notably, BMX with the LIME explainer and preselected parameters achieves an average improvement of \senew{0.087} points in 
\senew{Spearman} correlation on the system-level evaluation of SummEval.\footnote{We make our code available at: \url{https://github.com/Gringham/BMX}}
\end{abstract}

\input{sections/1_introduction}
\input{sections/2_background}
\input{sections/3_experiments}
\input{sections/4_results}

\input{sections/4_5_analysis}

\input{sections/5_related_work}
\input{sections/6_Conclusion}

\bibliographystyle{acl_natbib}
\bibliography{anthology,custom}

\appendix
\input{sections/9_appendix.tex}

\end{document}

%% file: sections/1_introduction.tex
\section{Introduction}
\label{sec:introduction}
Modern language model \rev{(LM)} based \rev{natural language generation} (NLG) metrics 
achieve astonishing results in grading machine generated sentences like humans would \cite[e.g.,][]{bhandari-etal-2020-evaluating,freitag-etal-2021-results,specia-etal-2021-findings,fabbri-etal-2021-summeval}. As most language models are black-box components, some recent works started to explore the explainability of \rev{LM-based} metrics \cite[e.g.][]{fomicheva-etal-2021-eval4nlp,leiter-explain,sai-etal-2021-perturbation,zerva-EtAl-2022-WMT, chen2023menli}. \rev{This exploration, for example, contributes to the foundation of ethical machine learning} \cite[e.g.][]{fort-couillault-2016-yes,eu-ethics}.

\begin{figure}[ht]
    \centering
    \includegraphics[width=0.49\textwidth]{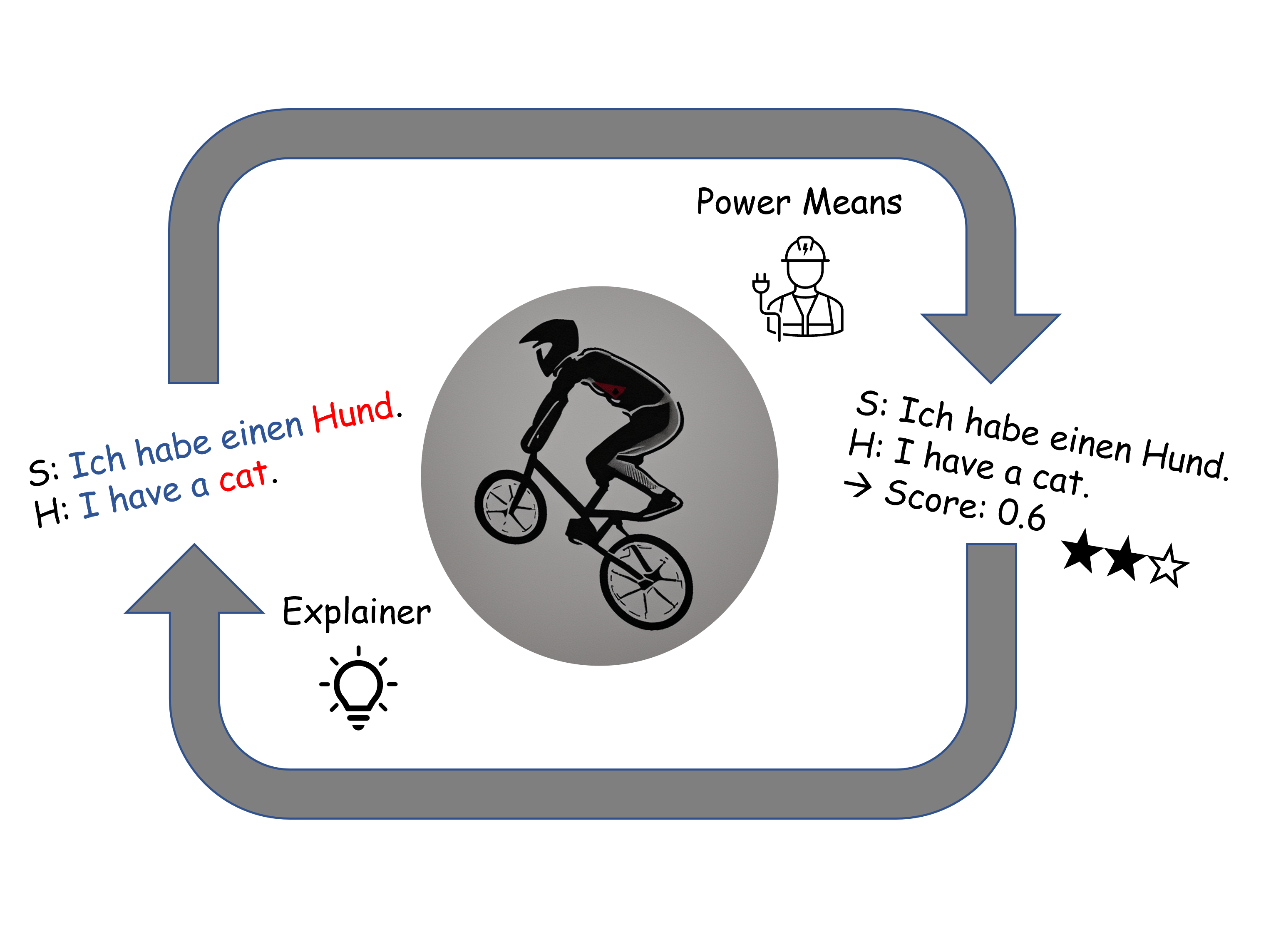}
    \caption{The duality of segment-level natural language generation evaluation metrics (right) and their word-level explanations (left).}
    \label{fig:bmx}
\end{figure}

\rev{Our work is motivated by an intriguing duality that we note between segment-level metrics and their explainability through feature importance techniques, e.g., LIME \cite{lime}:}\\
\textbf{Segment-level metrics}\footnote{We use the term \textit{segment-level}, as it includes the option that a metric grades multiple hypothesis sentences. \rev{Recent work shows that many sentence-level metrics also perform well on the segment-level \citep{deutsch-etal-2023-training}.}} return a single score indicating the quality of a generated \rev{segment}. \emph{Feature importance explanations}\footnote{Also called \textit{relevance scores} or \textit{attribution scores}.} increase the granularity of this score, by assigning additional \textbf{word-level scores}.  
 These granular scores capture additional information about the generated \rev{text} and about the metric that processed it, as, e.g., explored by the Eval4NLP21 shared task \cite{fomicheva-etal-2021-eval4nlp} and the WMT22 quality estimation shared task \cite{zerva-EtAl-2022-WMT}.
 \rev{On the other hand,} in recent \textit{multidimensional quality metrics} (MQM) datasets, word-level error annotations are converted into segment-level scores using heuristic functions \citep{freitag-etal-2021-experts}. Likewise, metrics like BERTScore \cite{bert-score} and BARTScore \cite{yuan-2021} build their segment-level scores upon word-level scores. In other words, we note the duality that feature importance techniques produce word-level scores from segment-level scores and heuristics can aggregate word-level scores into segment-level scores. \cli{Figure \ref{fig:bmx} gives an example \rev{of this duality} for machine translation (MT), where a German source sentence ``Ich habe einen Hund'' was wrongly translated into ``I have a cat''. On the right side, a segment-level score of 0.6 is assigned by a metric. On the left side, a feature importance explainer is used to explain this score by assigning word-level scores to each input token. Instead of displaying the scores, we use colors to describe the concept. The red words would likely achieve a low importance score, as they are translated incorrectly. The duality arises as the feature importance scores can be recombined into a new segment-level score (here using power-means).}

In this work, we explore whether this duality leads to iterative improvements of segment- and word-level scores, with a focus on segment-level scores as these are the main goal of modern metrics.
\rev{We propose \textit{\textbf{B}oosting natural language generation \textbf{M}etrics with e\textbf{X}plainability} (BMX)}, a method that directly leverages word-level explanations to improve the original segment-level score of a metric. Specifically, the approach aggregates word-level feature importance explanations using power means \cite{rueckle:2018} and combines them with the original score using a linear combination. To obtain the explanations, we leverage model-agnostic explainability techniques, allowing application to any NLG metric. \se{While we consider NLG (especially MT and summarization) as `natural use case', other regression and classification tasks follow similar settings, which makes our approach more generally applicable.} \cli{For example, in sentiment classification, feature importance techniques might assign high importance scores to tokens with positive sentiment. Hence, aggregating these scores could further inform a classification decision.}

We evaluate \rev{BMX} with several metrics and explainability techniques on 5 MT datasets \red{(3 for exploration + 2 held out for testing)}, as well as \senew{2} summarization datasets, and discuss conditions for its failure and success. Our work makes the following contributions:

\begin{itemize}[topsep=0pt,itemsep=-0.8ex,partopsep=1ex,parsep=1ex]
    \item[(i)] We highlight the duality of word-level explanations and segment-level scores for NLG metrics.
    \item[(ii)] We propose an approach to improve NLG metric\rev{s} by combining it with model-agnostic explainability techniques.
    \item[(iii)] We provide an evaluation that shows that our approach can achieve consistent improvements.
    \senew{For example, after applying BMX, we obtain 
    $0.087$ points improvement on SummEval.} 
\end{itemize}

%% file: sections/2_background.tex
\section{Approach}
\label{approach}

NLG metrics grade a generated \rev{text}, also referred to as hypothesis, by comparing it to a ground truth. For MT, 
the ground truth could be a human written reference translation or the original text in source language. For summarization, the ground truth could be a human written reference summary or the source text that is being summarized.
Given a pair of ground truth \rev{segment} $\boldsymbol{g}= \langle g_1, ..., g_n\rangle$ and hypothesis \rev{segment} $ \boldsymbol{h} =  \langle h_1, ..., h_m \rangle $, 
a segment-level metric $\mathcal{S}_0$ generates a single score $\mathcal{S}_0(\boldsymbol{g}, \boldsymbol{h})=\textit{s}_\textit{0} 
\in \mathbb{R}$. 
This score can be interpreted as, for example, 
the adequacy/accuracy of the generation of $\boldsymbol{h}$ given $\boldsymbol{g}$. 

Our algorithm consists of three steps: (1) compute feature importance explanations, (2) aggregate explanation scores, and (3) combine the aggregated explanations with the original score. 

\subsection{Feature importance computation}
The input of our algorithm is an arbitrary NLG metric $\mathcal{S}_0$, which we aim to improve, and a pair of ground truth and hypothesis \rev{segments} $(\boldsymbol{g}, \boldsymbol{h})$.  Further, we leverage a feature importance explainer $\smallE$, e.g., LIME \cite{lime} or SHAP \cite{DBLP:journals/corr/LundbergL17}. We use $\smallE$ to compute feature importance scores $\phi_i$ for each input token of an NLG metric. I.e., we explain $\mathcal{S}_0$ and its evaluation of $\boldsymbol{g}$ and $\boldsymbol{h}$ using $\smallE$ and obtain $\boldsymbol{\phi}\in\mathbb{R}^{m+n}$ as follows:
$$\smallE(\mathcal{S}_0,\boldsymbol{g}, \boldsymbol{h})=\langle \phi_1,...,\phi_n,\phi_{n+1},...,\phi_{n+m} \rangle$$

The importance scores $\boldsymbol{\phi}$ specify the contribution of each token in $\boldsymbol{g}$ and $\boldsymbol{h}$ to $\textit{s}_\textit{0}$. Note that the metric $\mathcal{S}_0$ itself is a parameter to $\smallE$ as model-agnostic explainers compare the metrics' original output with its output for permutations of the input \rev{text}. For a \rev{strong} metric, a high feature importance $\phi_i$ indicates that token $t_i\in\boldsymbol{g}\cup\boldsymbol{h}$ has a positive contribution to the score $\mathcal{S}_0$ and thus is likely to be correctly generated\footnote{\rev{For weak metrics, the segment-level score is incorrect more often, hence the feature importance scores are not as likely to be correlated to correct and incorrect translations.}}. Low feature importance can indicate incorrect translations or summaries. This setup follows the Eval4NLP21 shared task \cite{fomicheva-etal-2021-eval4nlp} for MT. 
Continuing the example from figure \ref{fig:bmx}, the source sentence ``Ich habe einen Hund'' is our $\boldsymbol{g}$ and the hypothesis sentence ``I have a cat'' is our $\boldsymbol{h}$; $\textit{s}_\textit{0}$ is $0.6$ and the output of $\smallE$ \rev{are} feature importance scores corresponding to the words, \se{e.g.} $\smallE(\mathcal{S}_0,\boldsymbol{g}, \boldsymbol{h})=\boldsymbol{\phi} = \langle0.5, 0.4, 0.2, 0.0, 0.5, 0.4, 0.2, 0.0\rangle$, where the low numbers indicate mistranslations. 

In some datasets, multiple reference\rev{s} are available for each hypothesis. In these cases, we concatenate the importance scores for each reference \rev{segment} into $\boldsymbol{\phi}$.

\subsection{Explanation score aggregation}
As mentioned above, the feature importance scores of a reasonable metric indicate the generation quality of each token. We combine these values to estimate the quality of the hypothesis at the segment-level. Therefore, we employ an aggregation function $f: \mathbb{R}^{m+n} \rightarrow \mathbb{R}$ to transform feature importance scores generated from the previous step into a single scalar value. We obtain the aggregated explanation score $\hat{s}_{0}$ as follows:
    $$f(\smallE(\mathcal{S}_0,\boldsymbol{g}, \boldsymbol{h}))) = \hat{s}_{0}$$
\subsection{Linear combination}
\label{linearcomb}
Finally, we \senew{linearly} combine $\hat{s}_{0}$ and $s_0$ using weight $w$ to construct a new metric $\mathcal{S}_1$:
    $$\mathcal{S}_1(\boldsymbol{g}, \boldsymbol{h})=w\cdot s_0 + (1-w)\cdot \hat{s}_{0} = s_1$$

We note that this three step process (feature importance computation, explanation score aggregation, linear combination) can be applied iteratively by increasing the index of $\mathcal{S}$ (resp.\ $s$). I.e., in the next iteration, we can consider $\mathcal{S}_1$ as the original metric and $s_1$ as the original score.

%% file: sections/3_experiments.tex
\section{Experiment Setup}
\label{expSetup}
In this section, we describe the datasets, metrics, explainers and aggregation methods that we evaluate in \S\ref{results} and their parameter configurations.

\subsection{Datasets}
\label{subsec:Datasets}
\cli{Our configuration of BMX has two parameters $w$ (see \S\ref{linearcomb}) and $p$ (see \S\ref{aggregation_technique}) which can either be selected \textit{in-domain} on a labeled subset of the same dataset or \textit{cross-domain} on a different dataset. We mainly evaluate cross-domain selection, as it would allow to apply BMX without additional annotation effort and is, therefore, more desirable. However, cross-domain tasks are generally also more difficult. For summarization, we also test an in-domain stratification approach. \rev{We refer to the datasets that we use for parameter search as \textit{calibration datasets} and to those that we evaluate on as \textit{evaluation datasets}.}}

\paragraph{MT datasets}
\cli{We use three \textit{calibration datasets}: the \textbf{WMT17} metrics shared task \cite{bojar-etal-2017-results} newstest2017 test set in the to-English direction, the 2020 partition of the \textbf{MLQE-PE} dataset \cite{fomicheva-etal-2022-mlqe} and the \textbf{Eval4NLP21} test set \cite{fomicheva-etal-2021-eval4nlp}. We evaluate BMX on two further \textit{evaluation datasets}: The \textbf{WMT22} Quality Estimation shared task \cite{zerva-EtAl-2022-WMT} and the \textbf{MQM}\footnote{We further refer to the datasets by these bolded names.} annotations of newstest21\footnote{https://github.com/google/wmt-mqm-human-evaluation} without human written references \cite{freitag-etal-2021-experts,freitag-etal-2021-results}. 
WMT17, MLQE-PE, Eval4NLP and WMT22 contain \textit{source sentence - hypothesis} pairs and human direct assessment (DA) scores \cite{graham_baldwin_moffat_zobel_2017} that grade the translation quality. For MLQE-PE, Eval4NLP21 and WMT22, human annotators determined these scores based on source and hypothesis sentences; for WMT17 they used reference sentences instead of source sentences. 
\rev{For \senew{MQM} \citep{lommel-2014}, \rev{scores}  are aggregated from fine-grained human MQM error annotations, and have been shown to be of better quality than crowdsourced annotations \cite{freitag-etal-2021-experts}.}
Table \ref{table:dataset_summary} \senew(appendix) shows an overview of the number of samples per language pair and dataset. }

\paragraph{Summarization datasets}
\cli{We perform \textit{in-domain calibration} on \textbf{SummEval} \cite{fabbri-etal-2021-summeval}. To do so, we apply cross-validation and split SummEval into eight non-overlapping \textit{configuration} (7 with 208 samples and 1 with 144) and \textit{evaluation} (7 with 1392 samples and 1 with 1456 samples) splits. Also, we make sure that no source text in the configuration set has another hypothesis in the corresponding test set. SummEval contains multiple expert-annotated discrete scores for coherence, consistency, fluency and relevance each and 11 reference summaries per hypothesis. We average the expert annotations for each score.}

\cli{Further, we use the parameter values obtained on SummEval and perform \textit{cross-domain calibration} on \textbf{RealSumm} \cite{bhandari-etal-2020-evaluating}. \rev{SummEval and RealSumm have the same data source, but different annotations and a different selection segments.}}

\subsection{Base metrics}
\label{sssec:num2}
\cli{We test BMX with the following metrics. }
\cli{\paragraph{Reference-based} For summarization, we test BMX with BERTScore \citep{bert-score} and BARTScore \citep{yuan-2021}.}
\cli{\paragraph{Reference-free} For MT, we test BMX with XBERTScore \citep{bert-score, song-etal-2021-sentsim, leiter-2021-reference}\footnote{We refer to BERTScore variants that use multilingual language models as XBERTScore.}, XLMR-SBERT \citep{reimers-gurevych-2020-making}, TransQuest \citep{ranasinghe-2021} and COMET \citep{rei-etal-2021-references}.}

\cli{We report the exact metric configurations in Appendix \ref{sec:appendix_metric_conf}.}

\subsection{Explanation techniques}
\label{sssec:num1}
We explore the effectiveness of three model-agnostic explainers: \cli{\textit{Erasure} \citep{Li2016UnderstandingNN}, \textit{LIME} \citep{lime} and \textit{SHAP} \citep{DBLP:journals/corr/LundbergL17}} For implementation details refer to appendix \ref{explainerAppendix}.

\textbf{Multiple references}: We handle the computation of the hypothesis and multiple references separately by fixing all but one during each application of the explainers and applying the explainer separately to each of them. E.g., if we have one hypothesis and 11 references and use LIME with 100 permutations, we will apply it 12 times, resulting in 1200 permutations in total.

\subsection{Aggregation technique}
\label{aggregation_technique}
Following \citet{rueckle:2018}, we use the power mean (or generalized mean) as a generalization over different means to aggregate token-level \rev{feature-importance} scores. 
The power mean of $n$ positive numbers $e_1,\ldots,e_n$ is computed as:
\[ M_p (e_1,...,e_n) = \left(\frac{1}{n} \sum_{k=1}^{n} e^p_k\right)^{\frac{1}{p}}\]  
Depending on $p$, the power mean takes on the value of specific means, e.g.\  $p=-1$ is the harmonic mean, $p=1$ is the arithmetic mean, and $p=-\infty$ resp.\ $p=+\infty$ is the minimum resp.\ maximum. 
We experiment with $p$-values between $[-30, 30]$ in $0.1$ steps. 
The token-level scores resulting from the explanation technique can be negative, 
which is problematic for power means, as these are defined on positive numbers only\footnote{Inserting negative numbers may lead to discontinuities or complex numbers.}.
To guarantee positive importance scores, whenever there is a negative importance score for a token, we add a regularization term to all importance scores of the current ground truth/hypothesis pair. This term is the absolute value of the smallest importance score assigned to any token of this pair. 
\senew{Additionally,}  
we \senew{generally} add a 
constant $1\mathrm{e}{-9}$ to each importance score to avoid issues with fluctuations around 0. 
\cl{Future work could explore further methods of aggregation such as different settings of the Kolmogorov mean \cite{doi:10.1080/00031305.2016.1148632}.}

\subsection{Evaluation}
To evaluate the BMX metrics, we calculate the correlation on datasets with human annotated scores. E.g., we can compute Pearson correlations per sample as follows:
\begin{equation}\text{Pearson}(\:H(\textit{LP}, D)\:, \: \mathcal{S}_1(\textit{LP}, D, \mathcal{S}_0, \smallE, w, p)\:)\label{pearsoneq}\end{equation}
Here, $H$ returns the set of human scores for language pair $\textit{LP}$ and dataset $D$. $\mathcal{S}_1$ returns the new metric scores, when our method is applied to $\textit{LP}$ and $D$. Its further parameters are the original metric $\mathcal{S}_0$, the explainer $\smallE$, the weight of the linear combination $w$ and the $p$ value of the power mean. On WMT22, we evaluate the segment-level Spearman correlation. On the MQM dataset, we evaluate segment- and system-level Kendall correlations. Further, for SummEval we evaluate the system-level Spearman and Kendall correlations. Finally, for RealSumm we report the segment-level Pearson and system-level Kendall correlations. With this setup, we follow the evaluation of the datasets' origin papers. An exception is the system-level evaluation of the MQM dataset, where we report the Kendall correlation per language pair as done by \citet{freitag-etal-2021-experts}.  

%% file: sections/4_results.tex
\section{Results}
\label{results}
In this section, we evaluate the effectiveness \cli{of} BMX by correlating the results with human judgments of \senew{MT} 
and summarization quality annotated in the datasets described in \S\ref{subsec:Datasets}. To start, we \rev{calibrate} the parameters $p$ and $w$. 

\paragraph{Calibrating $p$ and $w$} 
We perform a grid search \cli{on the calibration sets (see \S\ref{subsec:Datasets})} to determine the parameters $w$ and $p$ \cli{for our evaluation of BMX on the evaluation sets}. 

\cli{For $p$, we test $600$ equally spaced values in $[-30,+30]$ and for $w$, we test $6$ equally spaced values in $[0,1]$ (where $w=1$ reproduces the original score). This results in $3000$ BMX configurations (without $w=1$) for every metric-explainer combination. Next, we evaluate all $p$-$w$-metric-explainer combinations on the respective calibration set(s). Specifically, for the MT calibration sets we evaluate with segment-level Pearson correlation (see ~Eq.~\ref{pearsoneq}) for each language pair, and for summarization we evaluate with system-level Kendall correlation.}

\cli{For our evaluation, we select the median of the $p$ and $w$ values that led to any increase over the original correlation on the calibration set(s) for each metric-explainer combination.\footnote{We note that, as a benefit of BMX, not much data is used for the in-domain calibration on SummEval, as the calibration sets have small sizes of $\sim$200 samples.}}

Our approach of selecting $p$ and $w$ is rather simple. Future work might consider more sophisticated ways of optimization, such as considering the areas of highest increase in the grid search or even learning a model to set the parameters based on input \rev{segments}.

\begin{figure}[htb]
\includegraphics[width=0.44\textwidth]{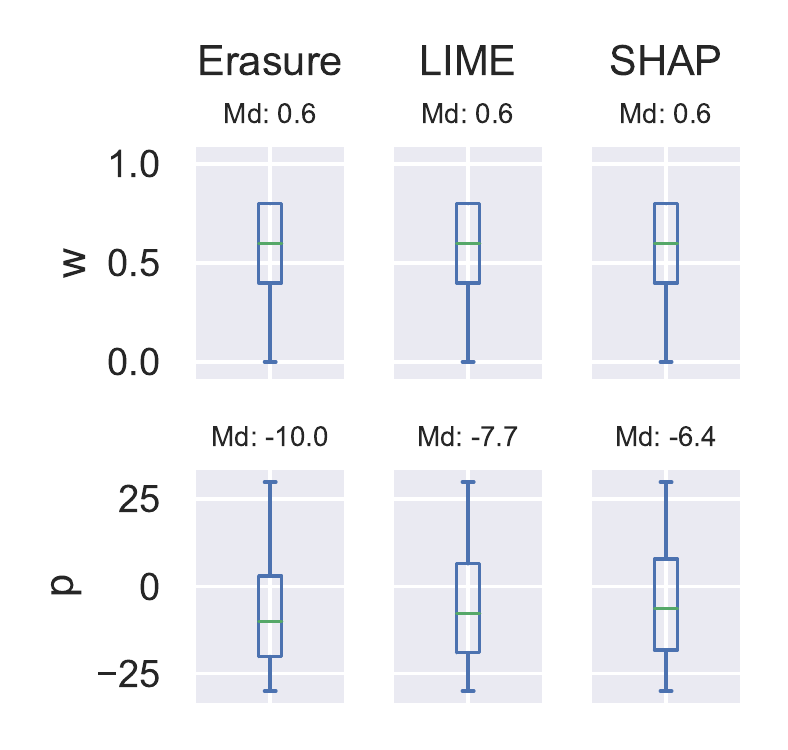}
\caption{Box-plots of the $w$ and $p$ values for XBERTScore leading to improvements with different explainers across all settings of the MT calibration sets. \textit{Md} denotes the Median value.}
\label{best_p_and_w}
\end{figure}

Figure \ref{best_p_and_w} shows exemplary box-plots of $p$ and $w$ for XBERTScore, to illustrate the distributions we select from.

\paragraph{Table structure} \label{tablest}
\cli{
In the next paragraphs, we present our results in Tables \ref{WMT22}, \ref{MQM}, \ref{SummEval1} and \ref{RealSumm}, using similar structures. The top row shows the metric names. For MT datasets, the left column shows the language-pairs. For SummEval, it describes the aspects graded by human annotators and whether Kendall (KD) or Spearman (SP) correlation is shown. For RealSumm, the left column describes whether segment-level Pearson or system-level Kendall evaluation is shown. Generally, the left-most number indicates the ORIGinal metric's correlation for each metric. The other numbers show the correlation of BMX using ERASure, LIME and/or SHAP respectively. 
Improvements over the original metric are colored in blue. For MT and RealSum, we print results in bold where improvements with BMX are statistically significant (p<=0.05) with the permute-both test described by \citet{deutsch-etal-2021-statistical}; underscored results remain significant after applying the Bonferroni-correction (per base metric; separately for MT and summarization) \citep{bonferroni, dror-etal-2017-replicability}.\footnote{We use the permute-both significance test implementation from \url{https://github.com/danieldeutsch/nlpstats} and the Bonferroni-correction implementation from \url{https://github.com/danieldeutsch/sacrerouge.}} \citet{dror-etal-2018-hitchhikers} describe that the statistical significance of cross-validation is underexplored. A simple solution they propose is to check that a predefined number of splits remains significant after applying the Bonferroni-correction. For SummEval, instead of selecting this predefined number, we report the number of significant splits.
Each average correlation we report has two superscript numbers. The first indicates the number of significant values before and the second after the Bonferroni correction (per base metric and correlation type). Results are rounded to 3 digits. Therefore, small improvements are indiscernable from the rounded numbers in some cases and can be identified by the coloring.}

\paragraph{Performance on WMT22}
\se{Table} \ref{WMT22} shows the performance with the preselected $p$ and $w$ values from the last section. 
BMX achieves an improvement in most cases, when running with XBERTScore and XLMR-SBERT, while it only improves TransQuest on two language-pairs. The average improvement with SHAP on XBERTScore and XLMR-SBERT is \senew{consistent but} rather small with 0.005 points in Spearman correlation. \cli{Notably, there are no improvements for the en-yo language pair of the WMT22 QE shared task \citep{zerva-EtAl-2022-WMT}. This language pair was introduced as a low-resource surprise set. The bad performance might be caused by the models not having seen much of Yoruba during training. Potentially BMX does not work here because there is nothing reasonable to explain, as the models do not know the language. }

\begin{table*} [htb]
\centering\small
\begin{tabular}{l|cccc}\toprule
LP & \textbf{XBERTScore} & \textbf{XLMR-SBERT} & \textbf{TransQuest} & \textbf{COMET}\\
& ORIG/ERAS/LIME/SHAP & ORIG/ERAS/LIME/SHAP & ORIG/LIME & ORIG/LIME
\\\midrule
en-cs & $0.294$/$\tablegreen{\textbf{0.295}}$/$\tablegreen{\underline{\textbf{0.314}}}$/$\tablegreen{\underline{\textbf{0.313}}}$ & $0.321$/$0.321$/$\tablegreen{0.327}$/$\tablegreen{\textbf{0.330}}$ & $0.556$/$0.545$ & $0.502$/$0.502$ \\
en-ja & $0.061$/$\tablegreen{0.062}$/$\tablegreen{0.064}$/$\tablegreen{\underline{\textbf{0.073}}}$ & $0.188$/$0.188$/$\tablegreen{0.189}$/$\tablegreen{\textbf{0.191}}$ & $0.275$/$\tablegreen{0.276}$ & $0.228$/$0.228$ \\
en-mr & $0.307$/$0.307$/$\tablegreen{\textbf{0.313}}$/$\tablegreen{\underline{\textbf{0.315}}}$ & $0.114$/$0.114$/$\tablegreen{0.115}$/$\tablegreen{\underline{\textbf{0.115}}}$ & $0.365$/$\tablegreen{0.367}$ & $0.291$/$0.291$ \\
en-yo & $-0.039$/$-0.039$/$-0.039$/$-0.040$ & $0.039$/$0.039$/$0.039$/$0.039$ & $0.066$/$0.066$ & $0.158$/$0.158$ \\
km-en & $0.569$/$0.569$/$\tablegreen{\textbf{0.573}}$/$\tablegreen{\underline{\textbf{0.575}}}$ & $0.477$/$0.477$/$0.477$/$\tablegreen{0.478}$ & $0.619$/$0.618$ & $0.443$/$\tablegreen{0.443}$ \\
ps-en & $0.558$/$0.558$/$\tablegreen{\textbf{\underline{0.562}}}$/$\tablegreen{\textbf{0.561}}$ & $0.446$/$0.446$/$0.446$/\tablegreen{\textbf{$0.446$}} & $0.614$/$0.614$ & $0.427$/$0.427$ \\
\midrule 
\text{AVG}
& $0.292$/$0.292$/$\tablegreen{0.298}$/$\tablegreen{0.299}$ & $0.264$/$0.264$/$\tablegreen{0.266}$/$\tablegreen{0.267}$ & $0.416$/$0.414$ & $0.342$/$\tablegreen{0.342}$\\
\bottomrule
\end{tabular}\caption{Segment-level Spearman correlation of metrics with and without BMX on the WMT22 dataset. We describe the table setup in the paragraph \textit{table structure} in section \ref{tablest}.}
\label{WMT22}\end{table*}

\paragraph{Performance on MQM}
Table \ref{MQM} shows the performance of BMX enhanced metrics for the MQM test set. On the segment-level, BMX improves all metrics in all language pairs, although only marginally for COMET. The average gain is $0.0075$ points in Kendall correlation.  In all but two cases, the improvement with BMX is significant. On the system-level BMX decreases the metric correlation for XBERTScore and Transquest. We investigate this in the paragraph \textit{MT failure analysis} in Section \ref{analysis} and find that better parameter selection can lead to strongly improved scores.

\begin{table*}[htb]
\centering\small
\begin{tabular}{l|cccc}\toprule
LP & \textbf{XBERTScore} & \textbf{XLMR-SBERT} & \textbf{TransQuest} & \textbf{COMET}\\& ORIG/LIME & ORIG/LIME& ORIG/LIME & ORIG/LIME
\\\midrule
en-de\_seg & $0.068$/$\tablegreen{\textbf{\underline{0.092}}}$ & $0.042$/$\tablegreen{\textbf{\underline{0.050}}}$ & $0.186$/$\tablegreen{0.188}$ & $0.248$/$\textbf{\underline{\tablegreen{0.248}}}$ \\
zh-en\_seg & $0.243$/$\tablegreen{\textbf{\underline{0.257}}}$ & $0.155$/$\tablegreen{\textbf{\underline{0.162}}}$ & $0.298$/$\tablegreen{\textbf{\underline{0.306}}}$ & $0.376$/$\tablegreen{0.376}$ \\
en-de\_sys & $0.051$/$0.051$ & $-0.051$/$-0.077$ & $0.245$/$0.231$ & $0.462$/$0.462$ \\
zh-en\_sys & $0.051$/$0.000$ & $0.103$/$0.103$ & $0.077$/$\tablegreen{0.103}$ & $0.564$/$0.564$ \\
\midrule 
\text{AVG}\_seg & $0.155$/$\tablegreen{0.174}$ & $0.099$/$\tablegreen{0.106}$ & $0.242$/$\tablegreen{0.247}$ & $0.312$/$\tablegreen{0.312}$\\
\text{AVG}\_sys & $0.051$/$0.025$ & $0.026$/$0.013$ & $0.161$/$\tablegreen{0.167}$ & $0.513$/$0.513$\\

\bottomrule
\end{tabular}
\caption{Segment- and system-level Kendall correlation of metrics with and without BMX on the MQM dataset. We describe the table setup in the paragraph \textit{table structure} in section \ref{tablest}.}\label{MQM}\end{table*}

\paragraph{Performance on SummEval}
Table \ref{SummEval1} shows the average Kendall and Spearman correlations of BMX (with in-domain calibration on the respective calibration splits) across the 8 test splits that we created from SummEval. In total, there is a \senew{strong} average gain of $0.074$ points in Kendall and $0.087$  points in \senew{system-level} Spearman correlation. 
\senew{Individually, gains are between 6-40\%, e.g., BERTScore improves from 0.309 to 0.431 Kendall.} 
These results show that, depending on the setting, BMX can substantially improve existing metrics.

\begin{table}[htb]
\centering\small
\begin{tabular}{l|cc}\toprule
Dataset & \textbf{BERTScore} & \textbf{BARTScore}
\\
& ORIG/LIME & ORIG/LIME     \\ \midrule
Coherence-KD & $0.533$/\tablegreen{$\textbf{0.675}^{5,4}$} & $0.202$/\tablegreen{$\textbf{0.229}^{2,2}$} \\
Consistency-KD & $0.029$/\tablegreen{$\textbf{0.142}^{4,4}$} & $0.513$/\tablegreen{$\textbf{0.519}^{0,0}$} \\
Fluency-KD & $0.294$/\tablegreen{$\textbf{0.356}^{4,1}$} & $0.420$/\tablegreen{$\textbf{0.448}^{2,0}$} \\
Relevance-KD & $0.379$/\tablegreen{$\textbf{0.550}^{8,8}$} & $0.415$/\tablegreen{$\textbf{0.458}^{5,2}$} \\
Coherence-SP & $0.690$/\tablegreen{$\textbf{0.831}^{8,8}$} & $0.289$/\tablegreen{$\textbf{0.324}^{3,1}$} \\
Consistency-SP & $0.022$/\tablegreen{$\textbf{0.211}^{6,6}$} & $0.708$/\tablegreen{$\textbf{0.723}^{1,0}$} \\
Fluency-SP & $0.389$/\tablegreen{$\textbf{0.467}^{5,4}$} & $0.389$/\tablegreen{$\textbf{0.467}^{2,1}$} \\
Relevance-SP & $0.465$/\tablegreen{$\textbf{0.608}^{8,8}$} & $0.555$/\tablegreen{$\textbf{0.601}^{5,2}$} \\
\midrule
\text{AVG-KD} 
&
$0.309$/\tablegreen{$\textbf{0.431}$} & $0.388$/\tablegreen{$\textbf{0.414}$}\\
\text{AVG-SP} & 
$0.391$/\tablegreen{$\textbf{0.529}$} & $0.528$/\tablegreen{$\textbf{0.563}$}\\
\bottomrule
\end{tabular}
\caption{Average system-level Kendall and Spearman correlation of metrics with and without BMX across the test splits we extracted from SummEval. We describe the table setup in the paragraph \textit{table structure} in section \ref{tablest}.}
\label{SummEval1}\end{table}

\paragraph{Performance on RealSumm}
Table \ref{RealSumm} shows the performance of BMX with BERTScore and BARTScore on the RealSumm dataset. We select the average of $p$ and $w$ values of the SummEval calibration splits for this setting as cross-domain calibration. BMX increases the system level correlation of BERTScore by $0.007$. However, for BARTScore the performance decreases.

\begin{table}[htb]
\centering\small
\begin{tabular}{l|cc}\toprule
Dataset & \textbf{BERTScore} & \textbf{BARTScore}
\\& ORIG/LIME & ORIG/LIME     \\\midrule
Segment & $0.304$/\tablegreen{$0.305$} & $0.488$/$0.474$ \\
System & $0.257$/\tablegreen{$0.264$} & $0.758$/$0.684$ \\
\bottomrule
\end{tabular}
\caption{Segment-level Pearson and system-level Kendall correlation of metrics with and without BMX for RealSumm. We describe the table setup in the paragraph \textit{table structure} in section \ref{tablest}.}
\label{RealSumm}\end{table}

%% file: sections/4_5_analysis.tex
\section{Analysis}\label{analysis}
\cln{In this section, we compare BMX to a fine-tuned metric on a SummEval split, analyze the failure in RealSumm and explore the stability of the metric when using the LIME explainer.}

\begin{figure}[htb]
\includegraphics[width=0.5\textwidth]{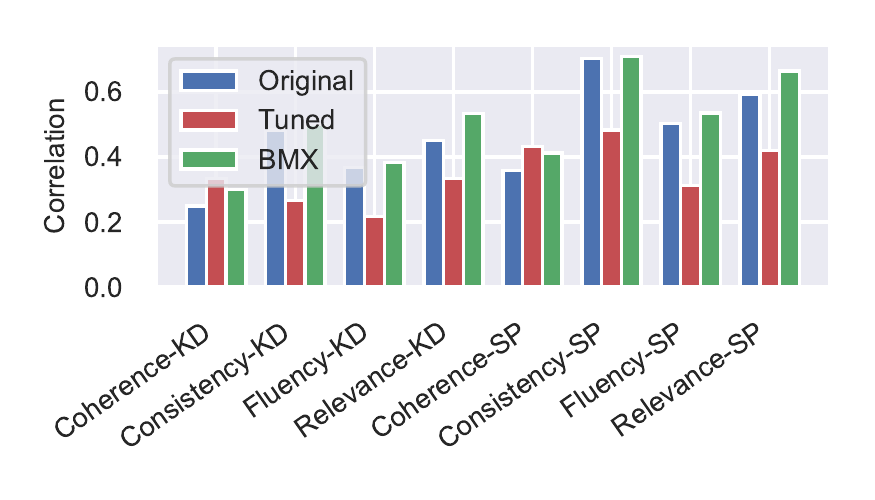}
\caption{System-level correlation with BARTScore on the first test split of SummEval. Left columns show the original correlation, middle columns show the correlation \cli{with BARTScore fine-tuned on the calibration set and right columns show the correlation with BMX.}}
\label{tune}
\end{figure}

\paragraph{Comparison to fine-tuning a metric}
We use the out-of-the-box training script of BARTScore to fine-tune BARTScore on the reference-hypothesis pairs of the first calibration split of SummEval. Then, we evaluate the fine-tuned metric, the original metric and BMX on the first test split and compare the results (see Figure \ref{tune}). The tuned metric has a better coherence than the original metric and BMX, however, all other aspects are worse than original. 
BMX has the highest correlation in all other dimensions, which shows that it can use the small-scale training set more efficiently.

\paragraph{MT failure analysis}
\cli{For some settings, for example with COMET, changes are extremely small. To understand BMX' internal workings, we plot the human scores and the two factors of the linear combination (the original score and the aggregated feature importance scores) for COMET on WMT22 cs-en (see the figure in Appendix \ref{cometdepp}). The scores are ordered by the human scores from high to low and normalized by z-scoring. We find that  \rev{many} scores that were aggregated from the explanations are uniform, with few outliers. Hence, adding them to the original COMET will hardly change the results. Future work could further explore the causes.}

As the system-level correlation decreased for some test setups on the MQM dataset, we further suspect that the transfer of $p$ and $w$ from the calibration sets to the evaluation set did not work out well, resulting in decreased correlations. To test this, we perform another grid-search on $p$ and $w$ and analyze whether other parameter settings would have performed better.  
The analysis shows that, even for COMET, the best parameter choice could lead to improvements of over $0.07$ Kendall points, with a choice of $w=0.2$ and a good selection of $p$ (see the figure in Appendix \ref{MQMDetail}). For Transquest, the improvements can be over $0.06$ Kendall points in en-de with $w=0.8$. 
Determining $p$ and $w$ in an in-domain setup might lead to better results. However, in real applications, there might not exist a human labeled portion of the dataset the method is applied to. Hence, future work could explore more elaborate mechanisms \cli{of selecting $p$ and $w$} than using the median of improvements on another dataset.

\begin{figure}[htb]
\includegraphics[width=0.5\textwidth]{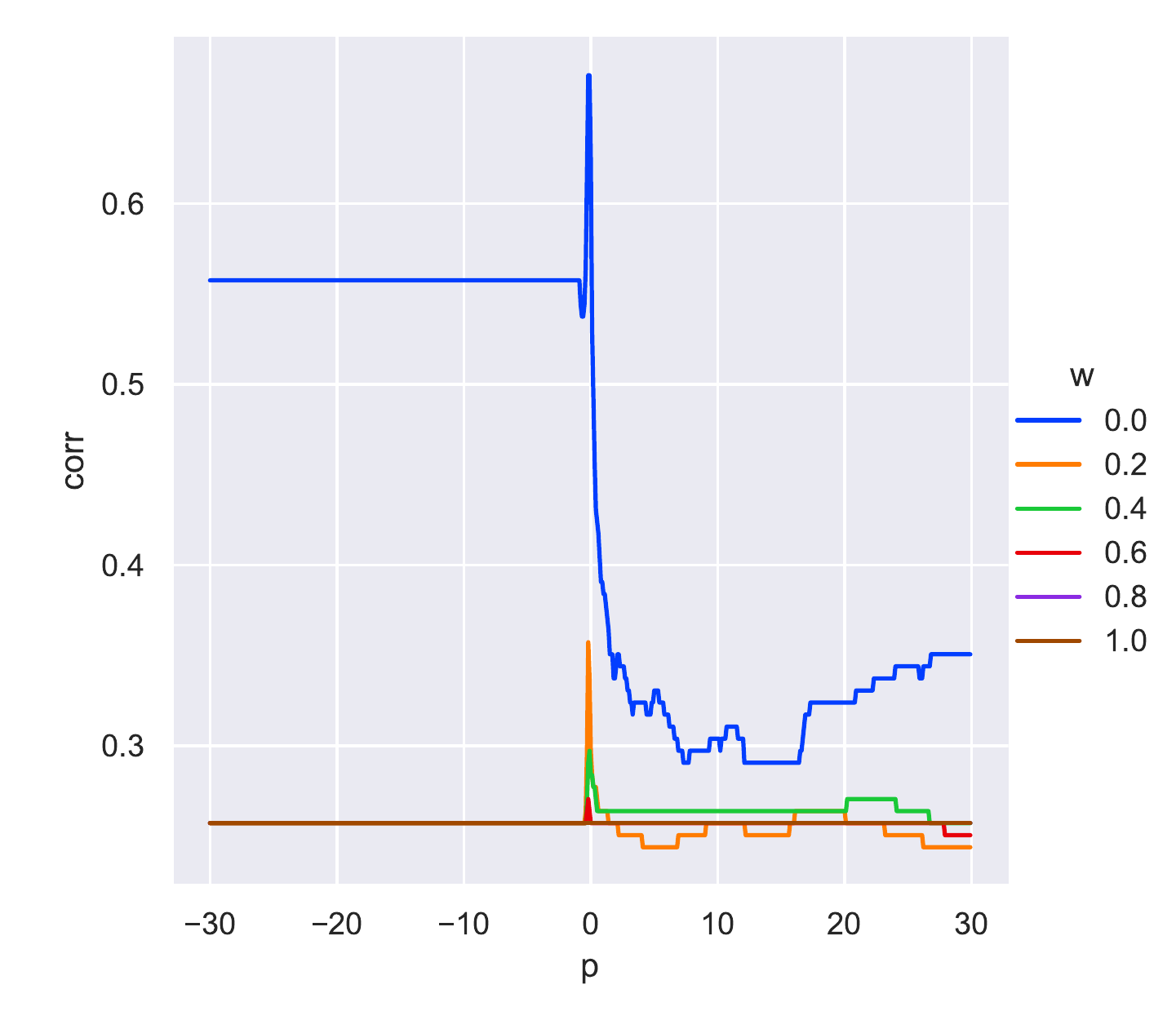}
\caption{System-level correlation with BERTScore on RealSumm, across $p$ values from $-30$ to $30$ and across $w$ values from $0$ to $1$, where $w=1$ is the original metric (indicated by a black line). BMX is using LIME in this sample.}
\label{RealSummDetail}
\end{figure}

\paragraph{RealSumm failure analysis}
We suspect that the transfer of $p$ and $w$ from SummEval to the domain of RealSumm did not work out well, resulting in decreased correlations. To test this, as for our MT failure analysis, we perform another grid-search on $p$ and $w$ and analyze whether other parameter settings would have performed better.  
The results of this analysis for BERTScore are visualized in figure \ref{RealSummDetail}. A choice of $w=0$ could have led to drastic improvements of over $0.3$ \senew{(over 100\% improvement)}. For BARTScore, the correlation could be improved by over $0.05$ with the correct selection (see appendix \ref{RealSummBartScoreDet}). Determining the values in a similar stratification setting as with SummEval might \senew{thus} have led to better results. 

\begin{figure}[htb]
\includegraphics[width=0.5\textwidth]{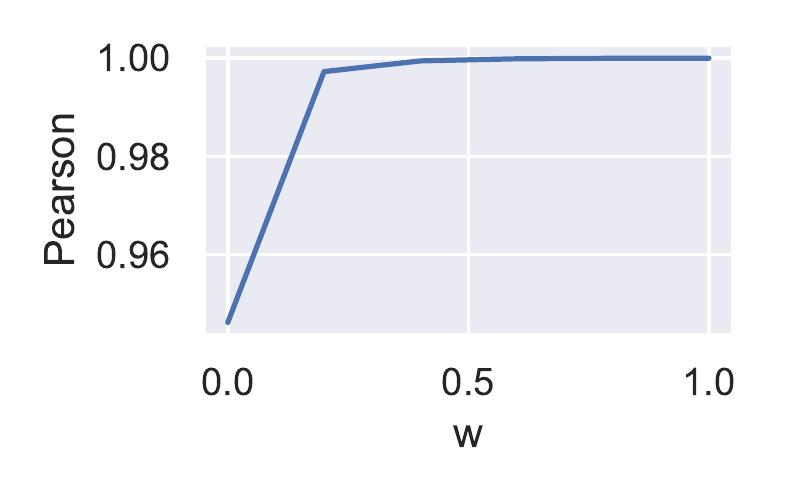}
\caption{Average Pearson correlation between 3 repeated runs of BMX with LIME and different settings of $w$ on the x-axis. The tests were computed on 3 language-pairs from WMT22 and the $p$-values range from -30 to 30 for every $w$ setting.}
\label{repeatability}
\end{figure}

\paragraph{Stability of LIME} As LIME uses random permutations, we test the stability of the approach for our task. To do so, we select the metric COMET and 3 language pairs of the WMT22 dataset. Then, we compute BMX with LIME using the grid-search configuration of the previous section. We exclude $w=1$, such that we get $3000$ scores per language pair. We repeat this process 3 times using $100$ permutations and 3 times using $1000$ permutations. Then we compute the average Pearson correlation among the first 3 runs and the last 3 runs. With $100$ permutations, the correlation is $0.9960$, indicating \senew{very} 
high stability of scores. With $1000$ permutations, it is $0.9997$. \senew{Thus,} further runtime can be traded for more stability. Lower $w$ values are less stable than higher ones (see figure \ref{repeatability}). The case of $w=0$ does not appear in our experiment calibrations and is therefore not applied on the test sets.

\paragraph{Influence of WMT2017} In contrast to newer datasets, the WMT17 dataset that we use for calibration is crowdsourced \citep{bojar-etal-2017-results}. Hence, we investigate its impact on the parameter calibration by removing it and rerunning the experiments. This marginally improves correlation on the test sets (up to 0.002). These results can be seen as a sign of the robustness of our parameter selection method, although it is not optimal performance-wise.

\paragraph{Segment- and System-level}
Generally, we note that the performance increases with BMX tend to be higher on system-level tasks, while they are more stable, but small, on the segment-level. As our analysis shows, the correct parameter selection is very important and can lead to high improvements, but also decreased correlation. Again, we note that future work could explore parameter selection, such as specifically choosing the parameters for each input, for example, by using a trained model.

%% file: sections/5_related_work.tex
\section{Related Work}
\label{related}
Our work is related to the domains of explainability and NLG metrics.

\paragraph{NLG metrics}
While embedding based metrics perform very well, their internal workings have become increasingly complex and cannot be easily understood by humans. The recent shared tasks Eval4NLP \citep{fomicheva-etal-2021-eval4nlp} and WMT22 QE \citep{zerva-EtAl-2022-WMT} explore the usage of explainability techniques for \senew{MT} 
to tackle this issue and provide word-level explanations for segment-level metrics. Motivated by their work, we also use word-level explanations, but additionally aggregate them to improve the original score.

Considering existing metrics, our work is especially related to word-level metrics and metrics that can be considered self-explaining. Word-level metrics like word-level TransQuest \cite{ranasinghe-2021} 
\senew{(in MT)}
are designed to assign translation quality scores to each word instead of the whole \rev{segment}. They can be considered as self-explaining, as they provide the same kind of explanations external explainers would provide \cite{leiter-explain}. Some existing segment-level metrics are self-explaining in this sense as well, as they use segment-level scores that are constructed from other word-level outputs. E.g., BERTScore is based on word-level cosine similarities of contextualized word-embeddings and BARTScore is based on word-level prediction probabilities of a BART model. We also use word-level scores to construct a new segment-level score. However, to the best of our knowledge, our method is the first to leverage model-agnostic explainabilitiy techniques to extract additional word-level information that is incorporated into the final metric. This has the benefit of being applicable to any segment-level NLG metric. 
\cli{BERTScore also has a configuration option to use tf-idf weighting on a token level. This is similar to feature importance explanations in the sense that both techniques assign “importance” scores to words. However, they describe different kinds of importance. Tf-idf weighting considers the general importance of words in a text. So these scores do not relate to “importance of the input to the output score” and potential errors considered by a metric.}
The Eval4NLP shared task showed that explanations from self-explaining methods tend to be stronger than model agnostic approaches \cite{fomicheva-etal-2021-eval4nlp}. Our method can provide another way to incorporate these word-level scores into the final prediction that might be explored by future work. 
Future work might also explore to use other model-specific explainers, e.g.\  gradient based or attention based methods \cite[e.g.][]{treviso-etal-2021-ist}.

Another topic related to explainable NLG metrics are fine-grained annotation schemes themselves. For example, the word-level scores annotated in the Eval4NLP shared task \cite{fomicheva-etal-2021-eval4nlp} or fine-grained error annotations like MQM \cite{lommel-2014} allow for human annotation of explanations that could for example be used to compare the word-level scores in our experiments to.  

Further, our approach is conceptually related to recent large language model (LLM) based approaches (released subsequently to our first Arxiv submission), where the LLMs iteratively explain and refine their own textual outputs \citep[e.g.][]{madaan2023selfrefine}. Also, further works on metrics have started to employ LLM generated textual error reports in metric heuristics \citep[e.g.][]{kocmi-federmann-2023-gemba, fernandes-etal-2023-devil}. We differ from these approaches by not relying on LLMs, and by using external explainers and feature-importance explanations. 

\paragraph{Explainability}
We leverage model-agnostic explainability techniques to collect word-level importance scores. There are many \cli{works} that give an overview on the topic of explainability, e.g., \citet{lipton-2016-mythos, BARREDOARRIETA202082}.

Specifically, we want to highlight the similarity of our approach to the concept of simulatability \cite[e.g.][]{hase-bansal-2020-evaluating}. Here, a machine or a human tester tries to reproduce an original model's output or solve an additional task, using the explanations they receive. We also utilize explanation outputs to accomplish a specific task. However, our focus is not to evaluate the performance of the explainers, but rather to use them to improve metrics for NLG. 

%% file: sections/6_Conclusion.tex
\section{Conclusion}
We have presented \textit{BMX: Boosting natural language generation Metrics with eXplainability}, a novel approach that leverages the duality of NLG metrics and feature importance explanations to boost the metrics' performance. BMX leverages model-agnostic explainability techniques, so that it can be applied to any NLG metric. Additionally, it requires no supervision once the initial parameters for $p$ and $w$ are set, which might benefit fully unsupervised \senew{or weakly supervised} approaches to inducing evaluation metrics 
\cite{belouadi-eger-2023-uscore}. 
\cln{Our tests show consistent improvements for multiple configurations on all tested datasets. 
Notably, we demonstrate strong improvements for summarization with 0.074 points in Kendall correlation on the system-level evaluation of SummEval, \cli{being significant on many test splits}. On RealSumm, BMX is not as strong\cli{, but} our analysis shows that a better choice of $p$ and $w$ could lead to strong improvements on this dataset as well.} 

To the best of our knowledge, our approach is the first to leverage the duality of segment-level MTE metrics and their feature-importance explanations directly and we believe that it can lead a step forward towards integrating metrics with explainability. 
Future work should also consider to which degree BMX can improve the explainability of metrics and apply our framework to other regression and classification tasks, beyond MT \senew{and summarization} metrics. \senew{Future work should also examine how to effectively leverage higher-level iterations.}

\section*{Acknowledgements}
The NLLG group gratefully acknowledges support from the 
Federal Ministry of Education and Research 
(BMBF) via the research grant ``Metrics4NLG'' and the German Research Foundation (DFG) via the Heisenberg Grant EG 375/5-1.

\section*{Ethical Considerations}
Our work might lead to the development of better natural language generation metrics. These metrics could be used to develop better generation systems. For these generation systems there is the risk of malicious usage, e.g., in the generation of hate speech or fake news. We think the benefit of these applications outweighs their misuse and note that our work is only considering their evaluation and hence does not carry a risk itself.

\section*{Limitations}
The post-hoc explainers that we use reevaluate permutations of the hypothesis and ground truth \rev{segments} by calling the original metric. This leads up to a few thousand executions depending on the configurations of LIME and SHAP \senew{(for Erasure, the number of executions depends on the input size, thus is much lower)}. 
We advise to test the runtime on a few samples and if necessary, adapt the configuration to use less permutations. 

Another limitation is that $p$ and $w$ need to be calibrated. The most promising approach to do this would be to evaluate a labeled subset of the dataset the metric should be applied on. If this is not feasible, existing datasets with human scores can be used for the calibration. Tuning these two parameters is little effort compared to the billions of parameters of modern LLMs, \senew{thus is comparatively efficient and applicable in small data scenarios}. 
Further, due to time constraints, we did not evaluate all metric-explainer combinations. Further analysis might thus show that other settings work even better. 
\cln{In \S\ref{related}, we discuss metrics that produce word-level scores or are self-explaining by default. While not applicable to all metrics, every metric that falls into one of these two groups has another option to compute explanations. As the Eval4NLP shared task showed, these tend to be stronger than model agnostic approaches \cite{fomicheva-etal-2021-eval4nlp}. Also, while not explicitly denoted as explanations, they are often already incorporated into the final score, e.g.\ for BERTScore or BARTScore. Here, we note that our method can provide another way of incorporating these word-level scores into the final prediction that might be explored by future work. 
Future work might also explore other model-specific explainers, e.g.\ gradient based or attention based methods \cite[e.g.][]{treviso-etal-2021-ist}.}  \cli{Lastly, while BMX can potentially be applied to other NLG tasks and other domains in general, we did not test it.}

%% file: sections/9_appendix.tex
\label{sec:appendix}

\section{Library Configurations}
\label{sec:appendix_metric_conf}
We use the following library and metric versions:
\begin{itemize} [topsep=0pt,itemsep=-0.8ex,partopsep=1ex,parsep=1ex,leftmargin=*]
    \item \textbf{LIME}: 0.2.0.1
    \item \textbf{SHAP}: 0.41.0
    \item \textbf{transformers}: 4.20.1, 4.24.0
    \item \textbf{BARTScore, Reference-Based}: bartscore: May 2022, facebook/bart-large-cnn + bart.pth (406,290,432 Parameters). BARTScore \cite{yuan-2021} returns the average generation probability of a sentence by a fine-tuned BART model as score. We use the \textit{ref$\rightarrow$hyp} generation direction of BARTScore, while the authors of BARTScore propose to use the \textit{src$\rightarrow$hyp} generation direction for SummEval \cite{yuan-2021}. We use \textit{ref$\rightarrow$hyp} as we want to leverage the large number of references in SummEval when applying BMX.
    \item \textbf{BERTScore, Reference-Based}: bertscore: 0.3.11; roberta-large (267,186,176 Parameters), No idf-weighting. BERTScore \cite{bert-score} computes a sentence score from the cosine similarity of contextualized word-embeddings between two input sentences.
    \item \textbf{COMET, Reference-Free}: comet: 1.1.3; wmt21-comet-qe-mqm (569330715 Parameters).  We use COMET-QE \cite{rei-etal-2021-references}, which uses a dual-encoder approach based on XMLR-models fine-tuned on human scores. \footnote{The stronger CometKiwi \cite{rei-etal-2022-cometkiwi} is not yet available at time of writing this paper.}
    \item \textbf{TransQuest, Reference-Free}: transquest: 1.1.1 TransQuest/monotransquest-da-multilingual; wmt21-comet-qe-mqm (560941057 Parameters). TransQuest \cite{ranasinghe-etal-2020-transquest} is a reference-free \se{trained} metric \cli{for} \senew{MT}, 
    which employs an XLMR model fine-tuned on human quality estimation scores that grade the hypothesis based on the source sentence. This model directly predicts a segment-level score as the output.
    \item \textbf{XBERTScore, Reference-Free}: bertscore: 0.3.11; joeddav/xlm-roberta-large-xnli (459,120,640 Parameters), No idf-weighting. \citet{leiter-2021-reference} empirically showed that among multiple XLM-RoBERTa \cite{conneau-etal-2020-unsupervised} model variants, one fine-tuned on a cross-lingual NLI dataset \textit{XNLI}\footnote{XNLI XLMR-Model:  \url{https://huggingface.co/joeddav/xlm-roberta-large-xnli}} \cite{conneau-etal-2018-xnli} achieves strong results on the Eval4NLP21 \cite{fomicheva-etal-2021-eval4nlp} dataset. 
    \item \textbf{XLMR-SBERT}: stsb-xlm-r-multilingual (278,043,648 Parameters). We use XLMR to compute multilingual sentence embeddings \cite{reimers-gurevych-2020-making}. Specifically, we use the cosine similarity of source and target embeddings as another segment-level metric.
\end{itemize}
For Erasure we use our own implementations. 

\section{Machine Translation Dataset Overview}
See Table \ref{table:dataset_summary}.
\begin{table*} [htb]
\begin{center}
\begin{tabular}{|l|c|c|c|c|c|}
\hline
      & \multicolumn{1}{l|}{WMT17} & \multicolumn{1}{l|}{Eval4NLP} & \multicolumn{1}{l|}{MLQE-PE} & \multicolumn{1}{l|}{WMT22} & \multicolumn{1}{l|}{MQM} \\ \hline
 LPs& cs-en                          &   \textbf{ro-en}                          &   \textbf{ro-en}                  & en-cs & en-de   \\ \hline
 & de-en                          &   \textbf{et-en}                            &   \textbf{et-en}                   & en-ja   & zh-en   \\ \hline
 & fi-en                          &   ru-de                            &   si-en                       & en-mr  &\\ \hline
 & lv-en                          &    de-zh                           &   ne-en                     & en-yo    &\\ \hline
 & \textbf{ru-en}                          &                               & \textbf{ru-en}               & km-en        &    \\ \hline
 & tr-en                          &                               &  en-zh                         & es-en & \\ \hline
 & zh-en                          &                               &  en-de                        & & \\\hline
Per LP &    560/(501)                        & 1000                             & 1000                               &        ca.1000 & 9002/10131   \\ \hline
Total & 3871      & 4000        &  7000    & 6000 & 19133\\\hline
\end{tabular}
\caption{\label{table:dataset_summary}Summary of the MT datasets we are using for exploration. We list the language pairs (LPs) in each set, the number of samples per pair and the total number of samples. The bold LPs occur in multiple datasets. \red{For zh-en some sentences in the dataset could not be loaded, hence this pair has only 501 samples.}}
\end{center}
\end{table*}

\section{Early results: selection of LIME}
\label{LimePrelim}
\begin{figure}[htb]
\includegraphics[width=0.49\textwidth]{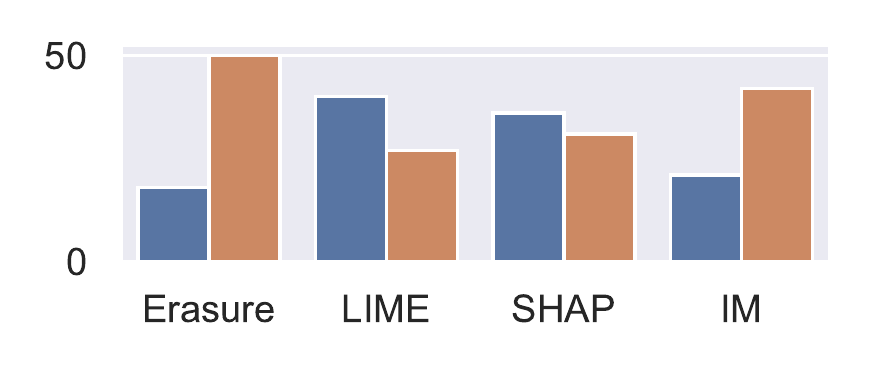}
\caption{Cases of improvement and decreased performance with $p$ and $w$ fixed to the respective explainer's best median. The blue bars show the number of settings with improved correlation, the orange bars show the number of settings with equal or worse correlation.}
\label{preselectedpw}
\end{figure}

We performed early experiments on WMT17, Eval4NLP and MLQE-PE, in which we selected the median of the p and w values that lead to the highest improvements per language-pair in a grid search. We only separated the values by explainer and not by metric. These experiments also included a variation of XMoverScore \cite{zhao-etal-2020-limitations} in the reference-free settings, as well as BERTScore and SentenceBLEU \cite{papineni-etal-2002-bleu} in the reference-based settings. XMoverScore is not included in the final experiments due to weak metric performance (we use it without target-side language model and cross-lingual mapping). BLEU and BERTScore are not included for machine translation, as only a few of the selected datasets provide reference sentences. It also included Input Marginalization \cite{kim-etal-2020-interpretation} as another explainer, which we didn't include in later experiments due to high runtime. Figure \ref{preselectedpw} shows a plot with the number of correlation improvements and decreases in each combination of language-pair, dataset and metric per explainer. We can see that LIME performs best, making it the default choice in the rest of our experiments.

\section{Implementation details for explainers}
\label{explainerAppendix}
\begin{itemize}
    \item \textbf{Erasure}: \citet{Li2016UnderstandingNN} suggest that model decisions can be investigated by analyzing the effect of feature removal. This is, e.g., used for adversarial attacks by \citet{li-2020}. We use Erasure to determine token-level importance scores by analyzing a metric's prediction with respect to the presence of each token in the translation. \red{I.e., for each token $t_i \in \boldsymbol{g}\cup\boldsymbol{h}$ we compute the importance $\phi_{i}$ as follows:
$$\phi_{i} = \mathcal{S}(\boldsymbol{g},\boldsymbol{h})-\mathcal{S}(\boldsymbol{g},\boldsymbol{h})_{/t_i}$$
where $\mathcal{S}(\boldsymbol{g},\boldsymbol{h})$ is an NLG metric grading the ground truth $\boldsymbol{g}$ and hypothesis $\boldsymbol{h}$. $\mathcal{S}(\boldsymbol{g},\boldsymbol{h})_{/t_i}$ denotes the same input without token $t_i$.}
    \item \textbf{LIME}: LIME \citep{lime} is a permutation based method, which trains a linear model that returns similar results as the explained model in a \textit{neighborhood} of inputs. Its weights are assigned to each corresponding word as feature importance explanations. When we explain a metric with LIME, for each ground truth or hypothesis sentence that is explained, LIME trains a linear model that returns similar results as the metric in a \textit{neighborhood} of this sentence. The \textit{dataset} used to fit this model is generated by randomly permuting the input.
    The labels of this dataset are determined by computing the metric score of \red{this permuted input}. Finally, the weights of the linear model are assigned to each token as feature importance explanations. We run LIME with 100 permutations per ground truth and per hypothesis sentence. We use the default replacement token of the LIME library \textit{UNKWORDZ}: \url{https://github.com/marcotcr/lime}. We use LIME with 100 permutations per hypothesis and ground truth each.
    \item \textbf{SHAP}: SHAP \citep{DBLP:journals/corr/LundbergL17} is an explainability technique that either exactly or approximately computes Shapley values from game theory, which measure the contribution of variables to a result, as feature importance scores. The exact SHAP explanation of a token is calculated using all possible permutations of the target sentence (with a single replacement token). The number of possible permutations grows exponentially with the number of input tokens. Therefore, SHAP is often approximated, e.g.\ using KernelShap \cite{DBLP:journals/corr/LundbergL17}. In our experiments, we use the same replacement string as for LIME: \textit{UNKWORDZ}.
    Also, up to a number of 7 tokens per sentence, we compute the exact SHAP. For more tokens, we use \textit{PermutationSHAP}, which is the default of the SHAP library\footnote{\url{https://github.com/slundberg/shap/blob/master/shap/explainers/_permutation.py}}.
\end{itemize}

\section{MQM with COMET}
\begin{figure}[htb]
\includegraphics[width=0.5\textwidth]{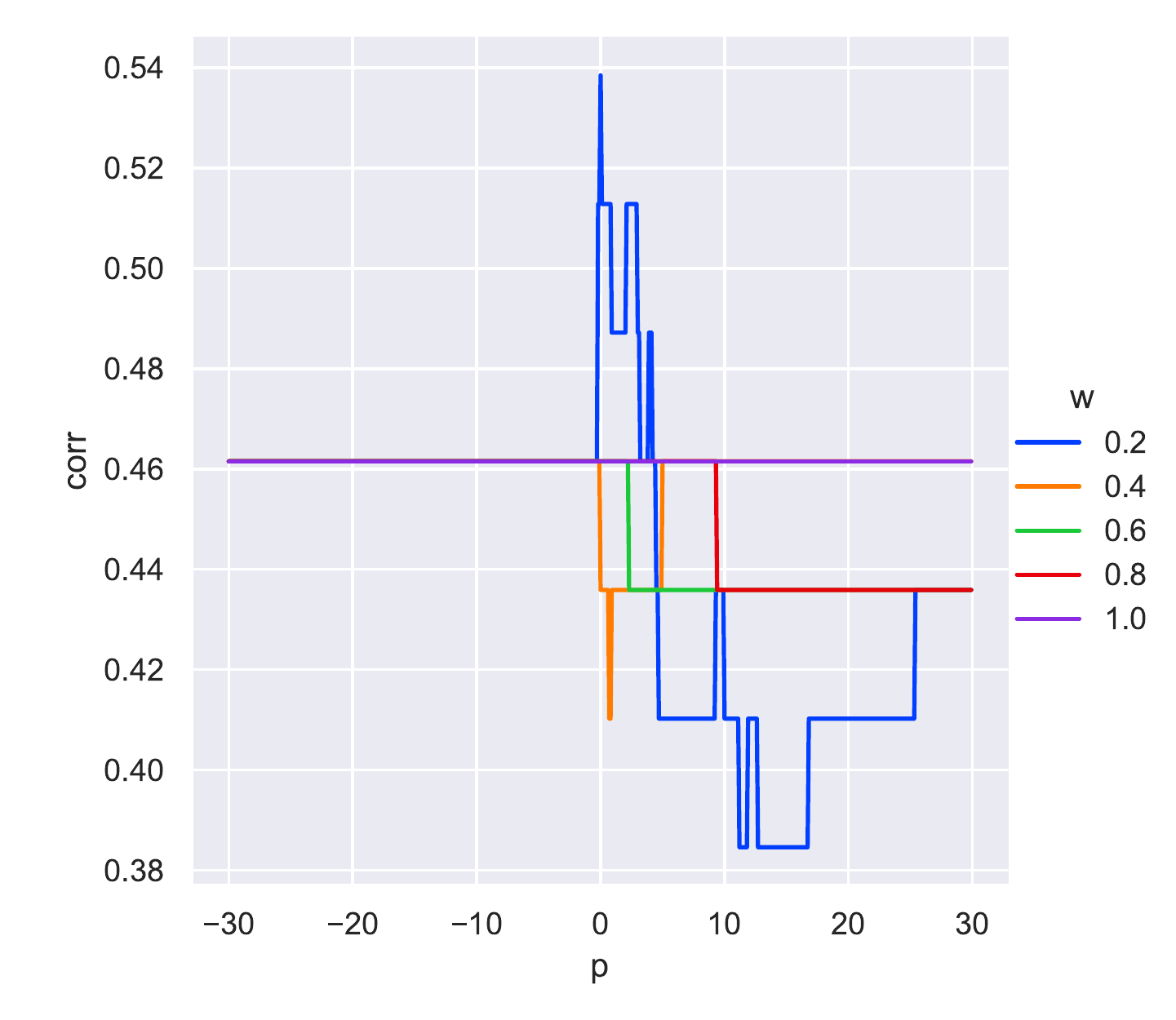}
\caption{System-level correlation with COMET on the MQM dataset, across $p$ values from $-30$ to $30$ and across $w$ values from $0$ to $1$, where $w=1$ is the original metric (indicated by a black line). BMX is using LIME in this sample.}
\label{MQMDetail}
\end{figure}

See Figure \ref{MQMDetail}.

\section{RealSumm with BARTScore}
\begin{figure}[htb]
\includegraphics[width=0.5\textwidth]{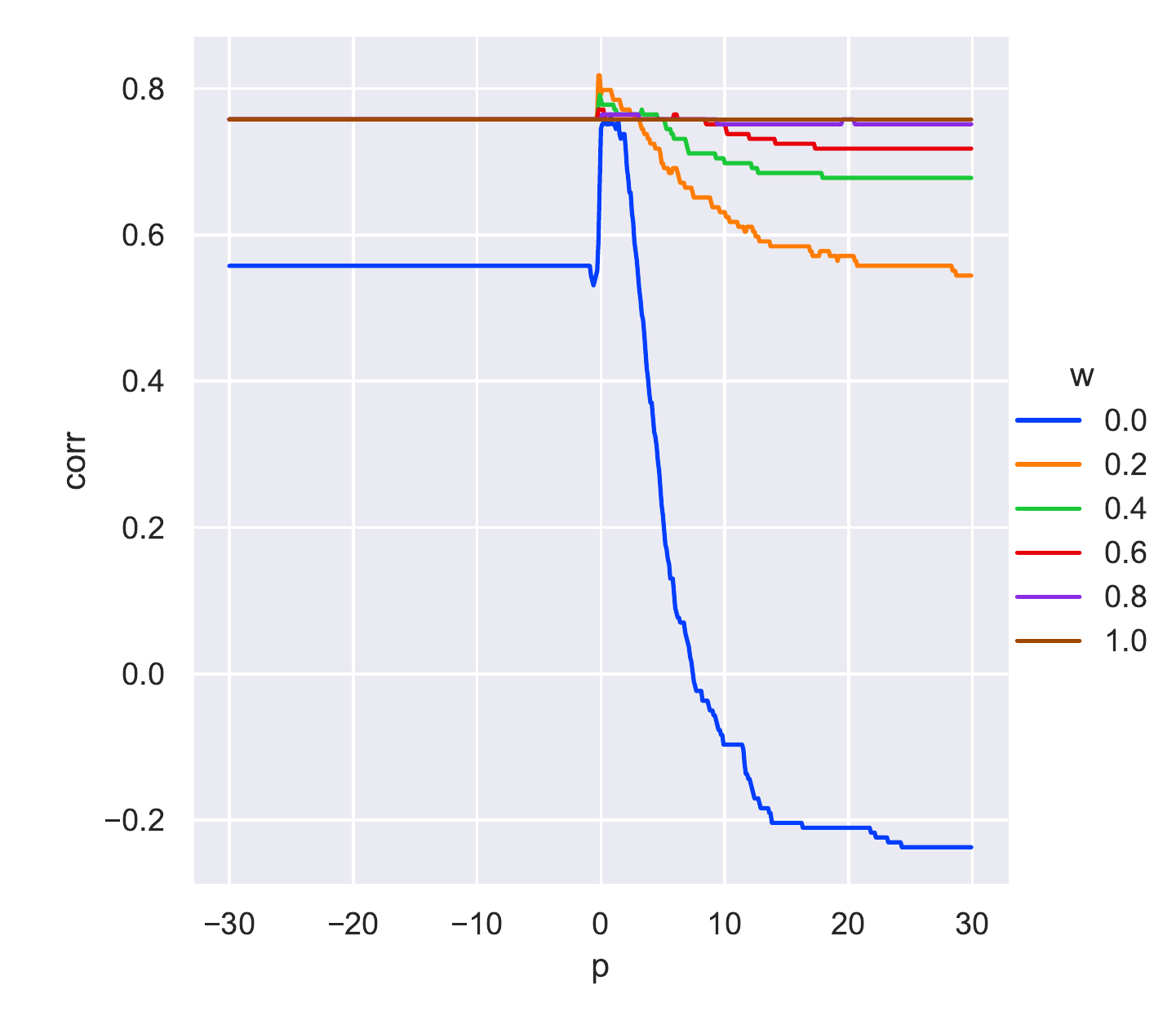}
\caption{System-level correlation with BERTScore on RealSumm, across $p$ values from $-30$ to $30$ and across $w$ values from $0$ to $1$, where $w=1$ is the original metric (indicated by a black line). BMX is using LIME in this sample.}
\label{RealSummDetailBARTScore}
\end{figure}

See Figure \ref{RealSummDetailBARTScore}.
\label{RealSummBartScoreDet}

\section{MT failure plot}
\begin{figure}[htb]
\includegraphics[width=0.40\textwidth]{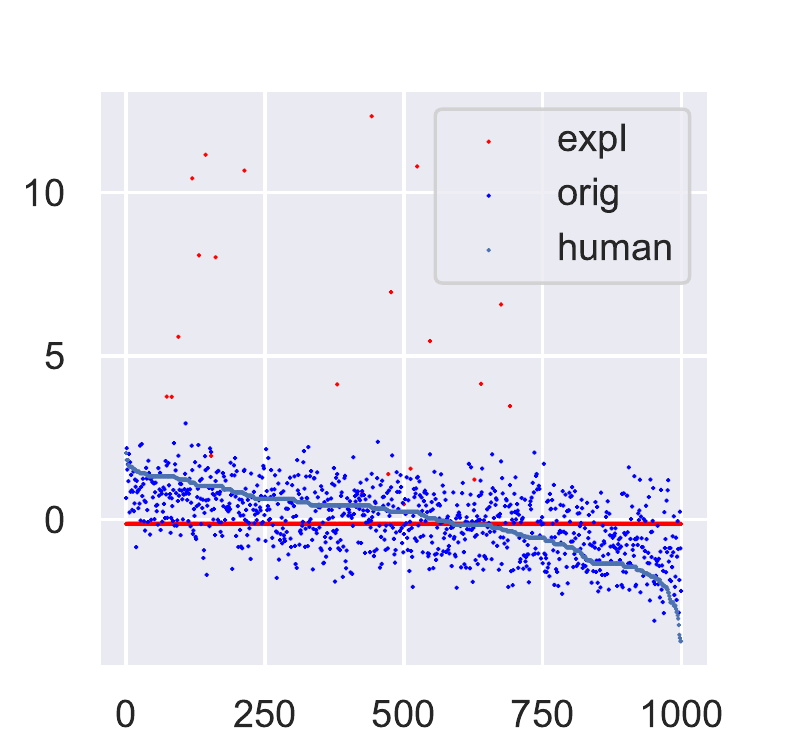}
\caption{Z-normalized original COMET scores, human scores and scores aggregated from explanations.}
\label{cometdep}
\end{figure}

See Figure \ref{cometdep}.
\label{cometdepp}
